\documentclass[10pt,twocolumn,letterpaper]{article}

\usepackage[pagenumbers]{wacv}      

\usepackage{graphicx}
\usepackage{amsmath}
\usepackage{amssymb}
\usepackage{amsfonts}
\usepackage{booktabs}
\usepackage{epsfig}
\usepackage{comment}
\usepackage{wrapfig}
\usepackage{microtype}
\usepackage{multirow}
\usepackage{soul}
\usepackage{enumitem}

\newcommand{\POISE}{\texttt{POISE}~}

\usepackage[pagebackref,breaklinks,colorlinks]{hyperref}

\usepackage[capitalize]{cleveref}
\crefname{section}{Sec.}{Secs.}
\Crefname{section}{Section}{Sections}
\Crefname{table}{Table}{Tables}
\crefname{table}{Tab.}{Tabs.}

\def\wacvPaperID{269} 
\def\confName{WACV}
\def\confYear{2024}

\begin{document}

\title{POISE: Pose Guided Human Silhouette Extraction under Occlusions}

\author{Arindam Dutta$^{1,* }$ \ Rohit Lal$^{1,* }$ \ Dripta S. Raychaudhuri$^{1,2, \dagger }$ \ Calvin-Khang Ta$^{1 }$ \ Amit K. Roy-Chowdhury$^{1 }$\\
$^{1}$University of California, Riverside $ \quad \quad ^{2}$AWS AI Labs \\
{\tt\small \{adutt020@, rlal011@, drayc001@, cta003@, amitrc@ece.\}ucr.edu} 
}
\maketitle
\newcommand\blfootnote[1]{%
  \begingroup
  \renewcommand\thefootnote{}\footnote{#1}%
  \addtocounter{footnote}{-1}%
  \endgroup
}
\blfootnote{* Equal contribution.}
\blfootnote{$\dagger$ Currently at AWS AI Labs. Work done while the author was at UCR.}

\maketitle

\begin{abstract}


Human silhouette extraction is a fundamental task in computer vision with applications in various downstream tasks. However, occlusions pose a significant challenge, leading to incomplete and distorted silhouettes. To address this challenge, we introduce \POISE: \ul{P}\ul{o}se Gu\ul{i}ded Human \ul{S}ilhouette \ul{E}xtraction under Occlusions, a novel self-supervised fusion framework that enhances accuracy and robustness in human silhouette prediction. By combining initial silhouette estimates from a segmentation model with human joint predictions from a 2D pose estimation model, \POISE leverages the complementary strengths of both approaches, effectively integrating precise body shape information and spatial information to tackle occlusions. Furthermore, the self-supervised nature of \POISE eliminates the need for costly annotations, making it scalable and practical. Extensive experimental results demonstrate its superiority in improving silhouette extraction under occlusions, with promising results in downstream tasks such as gait recognition. The code for our method is available \url{https://github.com/take2rohit/poise}.

\end{abstract}
\section{Introduction}
Human silhouette extraction is a fundamental task in computer vision with wide-ranging applications in human motion analysis \cite{poppe2007vision}, surveillance \cite{sharma2007extraction}, augmented reality \cite{adagolodjo2017silhouette}, and human-computer interaction \cite{chakraborty2018review}. The precise extraction of human silhouettes from images and videos enables the accomplishment of higher-level tasks, including gait recognition \cite{collins2002silhouette} and activity recognition \cite{kim2019vision}. However, the presence of occlusions \cite{chen2009frame} poses substantial challenges to the extraction process, resulting in incomplete or distorted silhouettes that impede the performance of downstream tasks.

\begin{figure}
    \centering
    \includegraphics[width=1\linewidth]{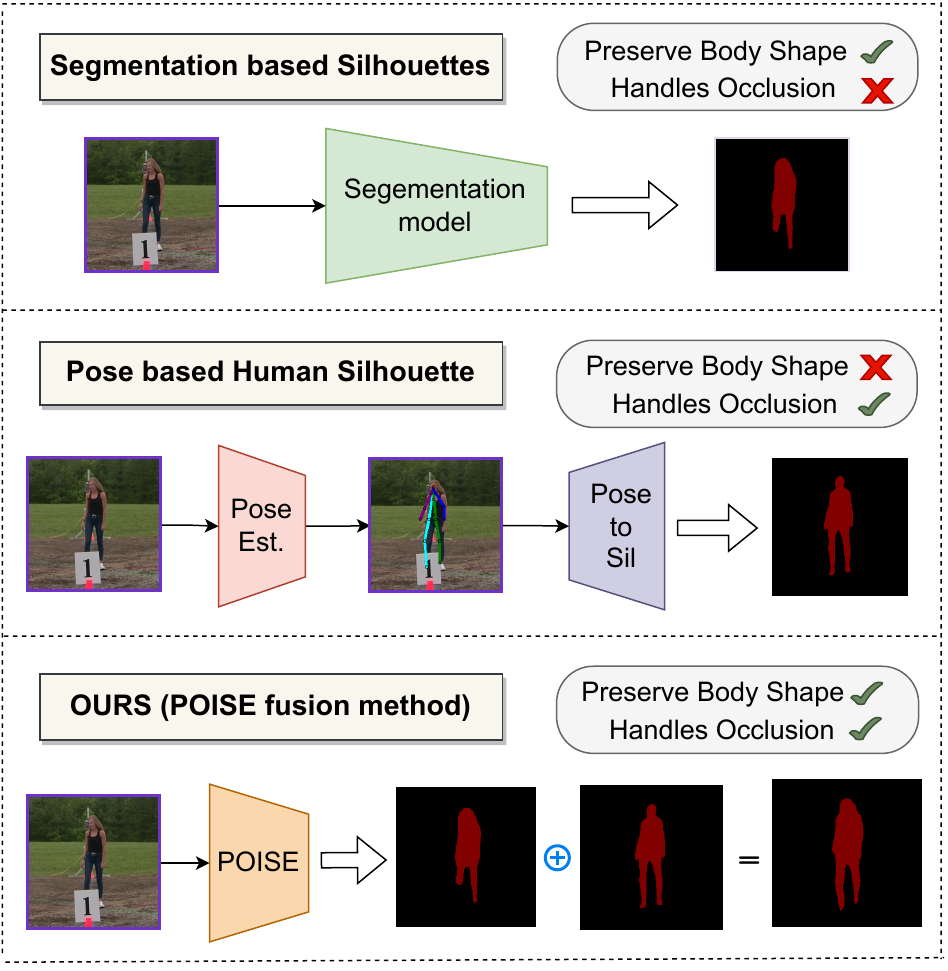}
    \caption{\textbf{Problem overview.} Existing works on segmentation struggle to effectively handle occlusion, leading to fragmented silhouettes (\textit{Top row}). One approach to remedy this is by using 2D pose estimates, however, this fails to  preserve the body shape (\textit{Middle row}). In this paper, we propose \POISE, a novel self-supervised framework that integrates both segmentation and 2D pose predictions to enable accurate prediction of human silhouettes under occlusion while preserving the body shape (\textit{Bottom row}).}
    \label{fig:teaser}
\end{figure}

Occlusions are common in real-world scenarios, where human subjects navigate complex and cluttered environments. They occur when certain areas of the human body are concealed or overlapped by objects or other individuals in the scene. Examples include objects obstructing body parts and self-occlusions caused by body parts coming into contact. State-of-the-art techniques for extracting human silhouettes, which involve pretrained semantic segmentation models like DeepLabv3 \cite{chen2017deeplab, chen2017rethinking}, struggle in such scenarios. Despite being trained on large-scale natural datasets, these models are not explicitly trained to handle occlusions. Consequently, they tend to misidentify occluded regions as background, resulting in fragmented and inaccurate silhouette predictions, as shown in Figure~\ref{fig:teaser}. Additionally, it is infeasible to acquire datasets with complete ground-truth human segmentation masks under occlusions due to visual uncertainty. 

To overcome these challenges, we present a novel fusion framework, \ul{Po}se Gu\ul{i}ded \ul{S}ilhouette \ul{E}xtraction under Occlusions (\POISE), that significantly enhances the accuracy and robustness of human silhouette prediction. Our approach combines the predictions from a segmentation model, which initially estimates the silhouettes, with the human joint predictions from a 2D human pose estimation model. The pose estimation model provides crucial spatial information by predicting the keypoints and inferring the body structure, aiding in handling occlusions. However, it lacks detailed information about the body shape. This is addressed via the initial silhouettes predicted by the segmentation model for the unoccluded regions, capturing precise body shape information. By fusing these two complementary sources of information, we refine the silhouette predictions, resulting in silhouettes that are unfragmented and preserve body shape, as illustrated in Figure~\ref{fig:teaser}.

Existing approaches for jointly learning poses and silhouettes primarily rely on supervised learning \cite{nie2018mutual, zhang2020correlating} which necessitates large annotated datasets for training. In this work, we relax this assumption and instead learn the fusion framework in an \emph{self-supervised fashion}. Specifically, given pretrained segmentation and pose estimation models, we train the fusion model to mimic the predictions given by the pair via a pseudo-labeling approach. In order to utilize the pose predictions for silhouette estimation, we design a novel auxiliary transformation function to convert the sparse keypoint estimates into dense human silhouettes. Our self-supervised approach alleviates the need for costly annotations and makes the method more scalable and applicable to real-world scenarios. \\

\noindent \textbf{Main contributions.} Our primary contributions are summarized as follows:
\begin{enumerate}
    \item We address the problem of self-supervised human silhouette extraction under occlusion.
    \item We present an effective fusion framework that leverages pretrained 2D pose estimation and silhouette extraction models to produce accurate and robust human silhouettes.
    \item Our framework is based on self-supervised learning, which eliminates the need for costly pixel-level annotations or pose annotations, thereby enhancing scalability.
    \item Beyond showcasing excellent results on silhouette extraction benchmarks, we also demonstrate the utility of the predicted silhouettes on downstream tasks such as gait recognition. 
\end{enumerate}

The rest of the paper is organized as follows: in Section \ref{sec:related_works}, we review prior works in human silhouette extraction, pose estimation, and gait recognition. Section \ref{sec:proposed_approach} describes the proposed framework in detail. Experimental results and analysis are presented in Section \ref{sec:experiments}, followed by conclusions and future directions in Section \ref{sec:conclusion_and_future_works}.

\section{Related Works}
\label{sec:related_works}

\noindent \textbf{Human Parsing and Silhouette Extraction.} Extensive research has been conducted on human silhouette extraction, with earlier works such as \cite{1699374} utilizing Hidden Markov models for modeling human silhouettes. More recent approaches leverage deep learning by primarily addressing the problem as semantic segmentation \cite{chen2017rethinking, long2015fully, chen2017deeplab, zhao2017pyramid, noh2015learning, ren2016faster,he2018mask, chen2020blendmask}. These methods have achieved high accuracy, but require large labeled datasets of human images with corresponding segmentation masks. Pretrained models like DeepLabV3 \cite{chen2017rethinking} perform well in diverse settings due to their training on extensive datasets. Despite the success of these methods, occlusion remains a challenge. Zhou {\it et. al.} \cite{zhou2021human} proposed a multistage architecture for de-occluding humans, while Li {\it et. al.} \cite{li2020self} introduced a state-of-the-art human parsing framework that can also be utilized for silhouette extraction by mapping all parts of the human body to the foreground. However, these algorithms require labeled training data, which is a limiting factor. 

\noindent \textbf{Human Pose Estimation.} Human pose estimation involves localizing keypoints on the human body, such as the head, elbows and knees, in 2D or 3D space. Deep learning-based pose estimation methods such as \cite{xiao2018simple, sun2019deep, martinez2017simple, zhang2019fast}, have achieved remarkable success on challenging academic datasets. However, these models are typically trained in supervised settings and often exhibit limited generalization capabilities when applied to unseen images. To overcome this, Zhang {\it et. al.} \cite{zhang2019unsupervised} proposed a novel domain adaptive 3D pose estimation algorithm. Jiang {\it et. al.} \cite{jiang2021regressive} proposed RegDA, a domain adaptive 2D pose estimation algorithm, which was further improved upon by Kim {\it et. al.} \cite{kim2022unified} in their work on UDAPE. Further, a source-free approach was recently proposed in \cite{raychaudhuri2023prior}. These unsupervised methods have made significant contributions to enhancing pose estimation in settings where labeled data is unavailable. 

\noindent \textbf{Human Pose Estimation under Occlusions:} Modern human pose estimation algorithms often fail to localize human keypoints under occlusions. To counter the same, Zhou {\it et. al.} \cite{zhou2020occlusion} introduced a novel algorithm based on siamese networks and feature matching to improve 2D human pose estimation performance under occlusions. Cheng {\it et. al.} \cite{cheng20203d} exploits spatio-temporal continuity for handing occlusions thus leading to improved pose estimation. Qiu {\it et. al.} \cite{qiu2020peeking} used a novel graph formulation for improving human pose estimation under multi-person occlusion scenario. Liu {\it et. al.} \cite{liu2022explicit} introduced a novel multi-stage framework for obtaining human keypoints under occlusions. Note that, these algorithms are entirely supervised learning algorithms necessitating costly annotations. To mitigate the same, Wang {\it et. al.} \cite{wang2022ocr} introduced a novel contrastive learning based occlusion-aware algorithm for predicting 3D human pose from given 2D keypoints. However, this algorithm still needs access to ground-truth 2D keypoints. 

\noindent \textbf{Multi-Task Learning for Pose and Parsing.} Multitask learning for human pose and body parsing aims to leverage the complementary information from both tasks to enhance their individual performances. Nie {\it et. al.} \cite{nie2018mutual} introduced an adaptive convolutional architecture that facilitates joint human pose estimation and body parsing. Their approach was trained in a supervised manner, minimizing a linear combination of losses for each task. Liang {\it et. al.} \cite{liang2018look} proposed a novel architecture that utilizes the semantic correlation between pose estimation and body parsing tasks. By exploiting this correlation, they were able to improve the accuracy of both tasks. More recently, Zhang {\it et. al.} \cite{zhang2020correlating} explored the explicit cross-task consistency between pose estimation and parsing. However, all of these methods are fully supervised and thus, require labeled data for both tasks. 

\noindent \textbf{Gait Recognition.} Gait recognition, a fundamental task in computer vision, involves identifying individuals based on their unique walking patterns. Silhouettes have been extensively studied and utilized for gait recognition for nearly two decades. Pioneering works by Collins {\it et. al.} \cite{collins2002silhouette} and Wang {\it et. al.} \cite{wang2003silhouette} laid the groundwork for employing silhouettes in this context. Since then, numerous studies \cite{wan2018survey} have explored the use of silhouettes for gait recognition. As gait recognition under occlusions poses a challenge, it is crucial to preprocess silhouettes to handle occlusions before incorporating them into gait recognition systems. This preprocessing step ensures the robustness and reliability of the gait recognition process, even in the presence of occlusions.

\begin{figure*}[!htb]
    \centering
    \includegraphics[width=0.75\textwidth]{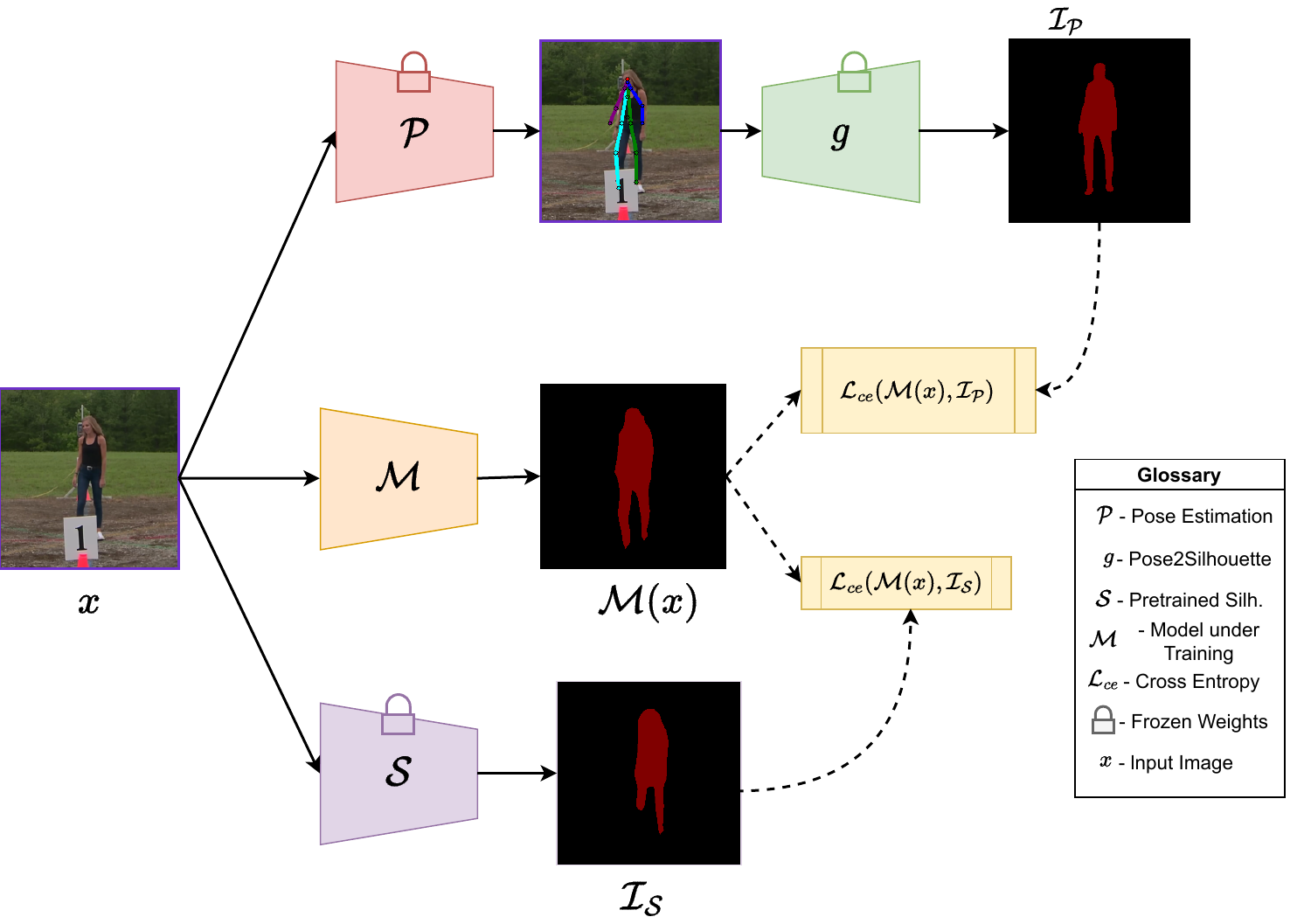}
    \caption{\textbf{Framework overview.} The key idea in \POISE is to use two noisy silhouettes: $\mathcal{I}_{S}$ and $\mathcal{I}_{P}$, obtained from $\mathcal{S}$ and $g \circ \mathcal{P}$ respectively, to train our model $\mathcal{M}$ to predict robust silhouettes under occlusion. Note that, $\mathcal{I}_{S}$ and $\mathcal{I}_{P}$ provide complimentary information - $\mathcal{I}_{S}$ retains identity specific shape features while $\mathcal{I}_{P}$ provides a generic human silhouette for the pose keypoints. \POISE learns a self-supervised fusion of $\mathcal{I_{S}}$ and $\mathcal{I_{P}}$.} 
    \label{fig:poise}
\end{figure*}

\section{Method}
\label{sec:proposed_approach}

\subsection{Problem Formulation}
Given a dataset of images $\mathcal{T}=\{x_i\}_{i=1}^{\textrm{N}}$, where each $x_i \in \mathbb{R}^{\textrm{H} \times \textrm{W} \times 3}$, our goal is to train a model $\mathcal{M}:\mathbb{R}^{\textrm{H}\times \textrm{W} \times 3} \rightarrow \mathbb{R}^{\textrm{H}\times \textrm{W}}$ that accurately predicts the silhouette $y$ corresponding to an input image $x$. To achieve this, we leverage the predictions of pretrained segmentation ($\mathcal{S}$) and 2D pose estimation ($\mathcal{P}$) models. These models provide initial predictions for the silhouette and sparse keypoints, respectively.

In order to align the sparse keypoints from $\mathcal{P}$ with dense human silhouettes, we utilize a pose-to-silhouette transformation function $g:\mathbb{R}^{\textrm{K}\times 2}\rightarrow \mathbb{R}^{\textrm{H}\times \textrm{W}}$, where $\textrm{K}$ represents the number of keypoints. We term this function \textbf{Pose2Sil}. By carefully fusing the predictions from $\mathcal{S}$ and the transformed keypoints using $g$, \POISE aims to produce accurate and robust silhouettes.

\subsection{Pose Guided Silhouette Extraction}
For each image $x \in \mathcal{T}$, we generate two silhouettes, $\mathcal{I_{S}}$ and $\mathcal{I_{P}}$, using pretrained models $\mathcal{S}$ and $\mathcal{P}$, respectively. Specifically,
\begin{subequations}
\begin{align}
    \mathcal{I_{S}} &= \mathcal{S}(x) \\
    \mathcal{I_{P}} &= g \circ \mathcal{P}(x) \ .
\end{align}
\end{subequations}
Here, $\mathcal{I_{S}}$ represents the silhouette obtained from the pre-trained segmentation model, while $\mathcal{I_{P}}$ is derived from the sparse pose predictions transformed by $g$. 

These pair of silhouettes provide complementary information. While $\mathcal{I_{S}}$ retains image-specific body shape details, it may become fragmented in the presence of occlusions. On the other hand, $\mathcal{I_{P}}$ produces a continuous silhouette but may lose accuracy in capturing the body shape. Thus, our problem of obtaining robust silhouettes under occlusions boils down to learning a mixed representation of the two individually noisy silhouettes. To learn a robust silhouette representation, we train $\mathcal{M}$ to effectively combine the information from these two noisy silhouettes under occlusions. This is accomplished by minimizing the pixel-wise binary cross-entropy loss between the predicted silhouette and those obtained from the pretrained models,
\begin{subequations}
\begin{align}
    \mathcal{L_{S}} &= \mathcal{L}_{\textrm{ce}}\left(\mathcal{M}(x), \mathcal{I_{S}}\right) \\
    \mathcal{L_{P}} &= \mathcal{L}_{\textrm{ce}}\left(\mathcal{M}(x), \mathcal{I_{P}}\right) \ . 
\end{align}
\end{subequations} 
Notably, we train our model to focus exclusively on the foreground information from $\mathcal{I_{S}}$ and ignore the background. This selective attention allows $\mathcal{M}$ to concentrate solely on the identity-specific features derived from $\mathcal{I_{S}}$. This is particularly crucial since the segmentation model may misidentify body parts as background due to occlusion, rather than the other way around. In practice, this is carried out via a simple masking operation, which ignores all background pixels in $\mathcal{I_{S}}$.

We also use the pseudo-labels obtained from $\mathcal{M}$ itself to regularize the training. Given an image $x$, we first obtain the pseudo-label $\mathcal{I}_{pl}=\mathcal{M}(x)$. These pseudo-labels are subsequently used to train the model using the pixel-wise binary cross-entropy loss,
\begin{align}
    \mathcal{L}_{pl} = \mathcal{L}_{\textrm{ce}}\left(\mathcal{M}(x), \mathcal{I}_{pl}\right) \ .
    \label{eqn:pl-loss}
\end{align}
In order to prevent noisy predictions from impeding the learning process, we use a confidence threshold $\tau$ to mask out possible incorrect pseudo-labels. This ensures only the high-quality predictions are reinforced by the model.

The overall training objective is given by
\begin{equation}
    \min_{\mathcal{M}} \lambda_{1}\mathcal{L_{S}} +  \lambda_{2}\mathcal{L_{P}} + \lambda_{3}\mathcal{L}_{pl} \ ,
    \label{eqn:trn-obj}
\end{equation} 
where each $\lambda_{i}$, for $i \in [1,3]$, controls the influence of individual loss terms in generating the final silhouette. An overview of our framework can be found in Figure~\ref{fig:poise}.

\begin{figure}[!htb]
    \centering
    \includegraphics[width=1\columnwidth]{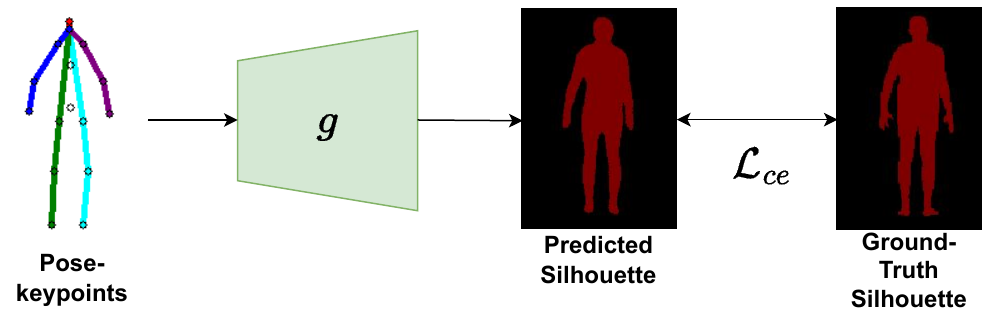}
    \caption{ \textbf{Learning Pose2Sil.} The model $g$ takes in the pose-keypoints: $p \in \mathbb{R}^{K \times 2}$ as inputs and predicts corresponding silhouettes: $s \in \mathbb{R}^{\textrm{H} \times \textrm{W}}$. The model $g$ is trained on a synthetic dataset $\mathcal{D}$ and is trained by minimizing the binary cross-entropy loss between the predicted silhouettes and ground-truth silhouettes. }
    \label{fig:pose2sil}
\end{figure}

\subsection{Obtaining Silhouette from Pose Keypoints}
\label{pose2sil}

To leverage the valuable information derived from the pose keypoints for training $\mathcal{M}$, we incorporate a dedicated module, \textbf{Pose2Sil}, which facilitates the transformation from pose to silhouette. In this section, we outline the training methodology employed for this module.

We assume access to an auxiliary \emph{synthetic} dataset $\mathcal{D} = \{(x_i,s_i,p_i)\}_{i=1}^\textrm{M}$ containing images of humans with corresponding annotations for pose keypoints $p$ and segmentation masks $s$. Due to its synthetic nature, it is simple to obtain annotations \cite{varol2017learning}, unlike the real-world images in $\mathcal{T}$. We train $g$ by minimizing the binary cross-entropy loss between the silhouette predicted from the pose and the corresponding ground-truth silhouette, $\mathcal{L}_{\textrm{ce}}\left(g(p),s \right)$. We illustrate this training process in Figure~\ref{fig:pose2sil}. Since the model $g$ does not rely on RGB images during training or inference, it remains unaffected by domain changes. Thus, it is domain-agnostic and can be applied seamlessly across different domains without retraining.

\subsection{Estimating Pose Keypoints under Occlusion}
\label{pose-est}

\begin{wrapfigure}{r}{0.25\textwidth}
    \centering
    \includegraphics[width=0.25\textwidth]{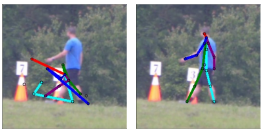}
  \caption{{\bf Limited generalization capacity of pre-trained pose estimation models.} \textit{Left:} Direct Inference of pre-trained model trained on dataset $\mathcal{D}$ on BRIAR dataset. \textit{Right:} Inference after adapting the pre-trained model to the same.}
  \label{fig:poseuda_example}
\end{wrapfigure}

Due to the limited generalization capabilities of pretrained pose estimation models \cite{jiang2021regressive, kim2022unified}, directly applying them to the images in $\mathcal{T}$ may result in subpar keypoint localization, as illustrated in figure~\ref{fig:poseuda_example}. Consequently, self-supervised domain adaptation of these pretrained models becomes imperative to ensure their effectiveness on our specific domain. 

In this work, we build upon the state-of-the-art domain adaptive pose estimation algorithm, UDAPE \cite{kim2022unified}. We leverage the availability of a synthetic dataset $\mathcal{D}$ as our source dataset, and $\mathcal{T}$ serves as our target dataset. The images in $\mathcal{T}$ may contain occlusions, which can significantly hinder the accuracy of pose estimation, as highlighted in previous research \cite{zhou2020occlusion}. To enhance pose estimation under occlusions on the target dataset, we introduce occlusions into the source dataset itself \cite{kundu2022uncertainty}. By doing so, we force the model to learn robust representations that facilitate improved pose estimation under occlusions in the target dataset. Please note that we do not have access to the ground-truth labels of $\mathcal{T}$.

The adaptation algorithm follows the self-training framework inspired by the widely-used Mean-Teacher approach \cite{tarvainen2017mean}. It involves the utilization of two identical models: a \emph{teacher} model and a \emph{student} model. Both models are initially initialized with the same weights at time step $t=0$. Subsequently, at each time step $t$, the student model parameters $\theta$ are updated by leveraging the supervisory signals provided by the teacher model, as well as the annotated data from the synthetic dataset $\mathcal{D}$. The parameters of the teacher model, denoted as $\Tilde{\theta}$, are updated using an exponential moving average (EMA) of the student model parameters, ensuring a smoother and more stable learning process. 

\begin{figure}[!htb]
    \centering
    \includegraphics[width=\columnwidth]{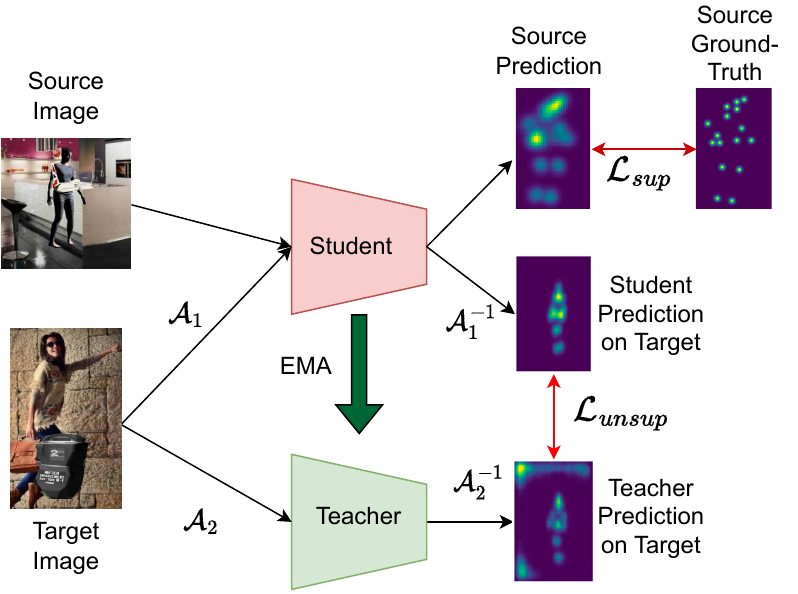}
  \caption{\textbf{Adaptation of the pose estimation model.} The algorithm is based on \cite{kim2022unified} with explicit occlusion handling by occlusion augmentation on the source dataset. $\mathcal{A}_{1}$ and $\mathcal{A}_{2}$ are augmentation operators such as rotation, scaling, etc. Similarly, $\mathcal{A}_{1}^{-1}$ and $\mathcal{A}_{2}^{-1}$ are inverse augmentation operators which essentially undo respective augmentations.}
  \label{fig:pose_adapt}
\end{figure}

\begin{figure}[!htb]
    \centering
    \includegraphics[width=0.75\columnwidth]{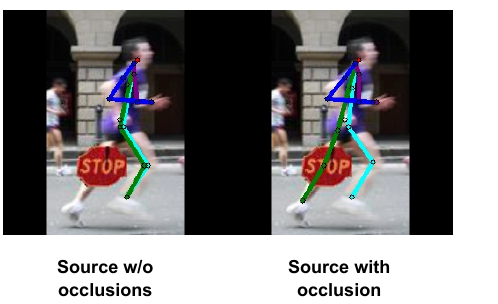}
    \caption{ \textbf{Domain Adaptive Pose Estimation under Occlusions.} \textit{Left:} Inference of adapted model \underline {without} occluding the source data ($\mathcal{D}$). \textit{Right:} Inference of adapted model \underline{with} occlusions on the source data ($\mathcal{D}$). }
    \label{fig:UDAPE-occl}
\end{figure}

To update the student model, a combination of supervised and self-supervised losses is employed. The supervised loss $\mathcal{L}_{\textrm{sup}}$ is computed using a mean square criterion on the source dataset, leveraging the annotated ground-truth labels. In addition to the supervised loss, a self-supervised consistency criterion $\mathcal{L}_{\textrm{unsup}}$ is introduced to encourage consistency in the pose predictions of two different augmentations of an image. Figure~\ref{fig:pose_adapt} shows a schematic diagram of the adaptation algorithm for learning $\mathcal{P}$. Once the adaptation is done, we use the updated teacher model as $\mathcal{P}$ to obtain the pose estimates on the images in $\mathcal{T}$. Figure~\ref{fig:UDAPE-occl} shows the importance of introducing occlusions in the source data which leads to improved pose estimation under occlusions.  

\section{Experiments and Results}\label{sec:experiments}

In this section, we provide a thorough assessment of \POISE, highlighting its exceptional ability to accurately extract human silhouettes even when they are partially obscured. We evaluate \POISE on five datasets, assessing its performance not only in human silhouette extraction but also in gait recognition. Our method outperforms existing off-the-shelf solutions and requires no extra annotations.

\subsection{Datasets}
We use the following datasets in our experiments.
\begin{itemize}[leftmargin=*,topsep=0pt]
\setlength\itemsep{-2pt}
    \item {\bf Humans3.6M} \cite{h36m_pami} is large scale real-world video dataset with over 3 million frames featuring from 11 professional actors performing different actions such as walking, eating, etc..  Following standard protocol, we use subjects 'S1', 'S5', 'S6', 'S7' and 'S8' for training and subjects 'S9' and 'S11' for testing. Similar to \cite{kim2022unified}, we use $\approx$ 20,000 frames for training and another $\approx$ 3000 for evaluation. We use the mean-IoU (mIoU) metric \cite{rezatofighi2019generalized} to report segmentation performance on this dataset.

    \item {\bf UP-S31} \cite{Lassner:UP:2017} is an image-based human part segmentation dataset with over 8000 images with 31 corresponding part annotations. However, several images containing more than one person were removed from our experiments leading to a total of 5226 images being used for our experiments. This was dis-jointly split into 4227 images for training and 1059 images for testing. We use the mean-IoU (mIoU) metric \cite{rezatofighi2019generalized} to report segmentation performance on this dataset.
    
    \item {\bf CASIA-B} \cite{yu2006framework} is a large-scale multi-view indoor gait recognition dataset with 124 subjects across 11 views. Each subject has a total 10 sequences - six sequences under Normal (NM) conditions and two each under Carrying Bag (BG) and Wearing different Clothing (CL) conditions. We adhere to existing works such as \cite{wu2016comprehensive} for training/testing and gallery/probe partitions. We report Rank-1 accuracy for gait recognition experiments using GaitBase \cite{fan2023opengait} on this dataset.
    
    \item {\bf BRIAR} \cite{cornett2023expanding} is a recent large-scale real-world biometric dataset with images of individuals under challenging conditions such as atmospheric turbulence and natural occlusions. For this work, we prepare the frames by tracking the person with Byte-Track \cite{zhang2022bytetrack}. We then select about 20,000 frames at different distances, allocating roughly 75\% for training and the remainder for testing. Since we lack ground-truth silhouettes, we present qualitative silhouette extraction results. Additionally, we evaluate gait recognition using GaitBase \cite{fan2023opengait} on a subset of 87 subjects divided into 60 for training and 27 for testing. 

    \item {\bf 3DOH50K} \cite{zhangoohcvpr20} is large-scale real-world occlusion dataset with over 50000 frames. As the ground-truth segmentation masks are incomplete and images do not have ids associated with them, we report qualitative results for silhouette extraction on this dataset. 


\end{itemize} 

\noindent \textbf{Generating occluded images.} 
We use the datasets mentioned above to generate corresponding occluded RGB images for evaluation purposes. In particular, we consider two kinds of occlusions - Random Erase (RE) Occlusions \cite{zhong2020random} and Common Objects in Context (COCO) Occlusions \cite{lin2014microsoft}. For both Random Erase and COCO Occlusions, we first estimate keypoints on clean images using the adapted pose estimation model described in Section \ref{pose-est} and then add occlusions on randomly selected keypoints. This ensures that the occlusion always covers a certain part of the human body. Additional details on the same is presented in the section 1 of the supplementary.

\subsection{Implementation details}
In our experiments, we employ the DeepLabv3 architecture \cite{chen2018encoder} with the ResNet-101 \cite{he2016deep} feature extractor as the backbone for our model $\mathcal{M}$. For most of our experiments, we leverage the ResNet-101 pre-trained on the COCO dataset \cite{lin2014microsoft} as the segmentation network $\mathcal{S}$. However, for the specific scenario of COCO occlusions on Huamns3.6M and UP-S31 dataset, we adopt the SCHP architecture \cite{li2020self} pre-trained on the LIP dataset \cite{gong2017look} as $\mathcal{S}$. For experiments with Random Erase Occlusions, $\mathcal{M}$ is trained with five different severities of occlusion (12, 16, 20, 24, and 28), and the same model is used for inference on the five different severities of occlusion. Data augmentation strategies such as random rotation, translation, and shear are also used to regularize the training process. Additional implementation details are provided in the section 2 of supplementary. 

\noindent \textbf{Synthetic Dataset.} We use SURREAL \cite{varol2017learning} as our synthetic dataset $\mathcal{D}$. SURREAL is a large-scale dataset containing over 6 million frames of synthetically generated human images against an indoor background.  

\noindent \textbf{Pose Estimation Model ($\mathcal{P}$).} Adhering to \cite{kim2022unified}, we use the Simple Baseline decoder \cite{xiao2018simple} with Resnet-101 backbone \cite{he2016deep} the architectures for our student and teacher networks. The networks are trained for a total of 80 epochs with the first 40 epochs being used for supervised training of the student model and the other 40 epochs for adaptation to the target domain. We use a batch size 32 while optimizing using the Adam optimizer \cite{kingma2014adam} with an initial learning rate of $1e-4$, decaying by a factor of $0.1$ after $45^{th}$ and $60^{th}$ epochs. 

\noindent \textbf{Pose2Sil Model ($g$).} The architecture of $g$ is the same as that of the DCGAN \cite{radford2015unsupervised}. The model is trained for a total of 200 epochs using the Adam optimizer \cite{kingma2014adam} with a learning rate of $1e-4$ and a batch size of 32.

\begin{table}[!htb]
\centering
\caption{Quantitative Results using mIoU metric for \POISE against $\mathcal{I_{S}}$ and $\mathcal{I_{P}}$ on the Humans3.6M dataset with COCO occlusions.}
\begin{tabular}{@{}ll@{}}
\toprule
Method & mIoU \\
\hline 

$\mathcal{I_{S}}$ & 80.14   \\
$\mathcal{I_{P}}$ & 75.97   \\   
\POISE & 87.22\\
\POISE + Weak Sup. & 89.67   \\
Full Sup. & 92.75   \\
\bottomrule
\end{tabular}
\label{tab:h36m-miou}
\end{table}

\begin{table}[t]
\centering
\caption{Quantitative Results using mIoU metric for \POISE against $\mathcal{I_{S}}$ and $\mathcal{I_{P}}$ on the UP-S31 dataset. RE - Random Erase Occlusion, COCO - COCO occlusion.}
\begin{tabular}{@{}llll@{}}
\toprule
\multirow{2}{*}{\begin{tabular}[c]{@{}l@{}}Occlusion\\ Severity\end{tabular}} & \multirow{2}{*}{ $\mathcal{I_{S}}$} & \multirow{2}{*}{$\mathcal{I_{P}}$} & \multirow{2}{*}{$\POISE$} \\
& & & \\ \midrule
RE - 12& 79.67                             & 81.98                             & \textbf{85.58}           \\
RE - 16                                                                       & 78.25                             & 81.72                             & \textbf{85.35}           \\
RE - 20                                                                       & 76.71                             & 81.43                             & \textbf{85.11}           \\
RE - 24                                                                       & 74.88                             & 81.34                             & \textbf{84.55}           \\
RE - 28                                                                       & 73.28                             & 80.84                             & \textbf{83.66}           \\
COCO                                                                          & 78.68                             & 77.34                             & \textbf{80.86}           \\ \bottomrule
\end{tabular}
\label{tab:ups31-miou}
\end{table}

\subsection{Results}

\subsubsection{Silhouette Extraction}
\label{robust-silh-ext}

We evaluate the effectiveness of \POISE for human silhouette extraction under occlusion in terms of segmentation accuracy. Table ~\ref{tab:h36m-miou} shows the efficacy of \POISE is learning strong feature representations to perform human silhouette extraction under COCO occlusions on Humans3.6M dataset. While our results are inferior to a fully-supervised baseline by $\approx$ 5.5 \%, we show that we can bridge this gap to $\approx$ 3 \% by fine-tuning the network with limited supervision, i.e. considering 5 \% of the training dataset as annotated. This shows that \POISE learns generalized feature representations that can be effectively used to obtain optimal human silhouettes under occlusions.  We report the mIoU on UP-S31 dataset at six different occlusion severities in Table~\ref{tab:ups31-miou}. We note that for Random Erase occlusions, \POISE provides an improvement of $\approx 8\%$ against $\mathcal{I_{S}}$ and $\approx 3\%$ against $\mathcal{I_{P}}$. For COCO occlusions, \POISE provides an improvement of $\approx 2\%$ against $\mathcal{I_{S}}$ and $\approx 3\%$ against $\mathcal{I_{P}}$. This shows the deficiencies of using state-of-the-art human segmentation methods in scenarios involving occlusion. We present additional qualitative results on the same in the section 3 of the supplementary.

\begin{figure}[t]
    \centering
    \includegraphics[scale = 0.45]{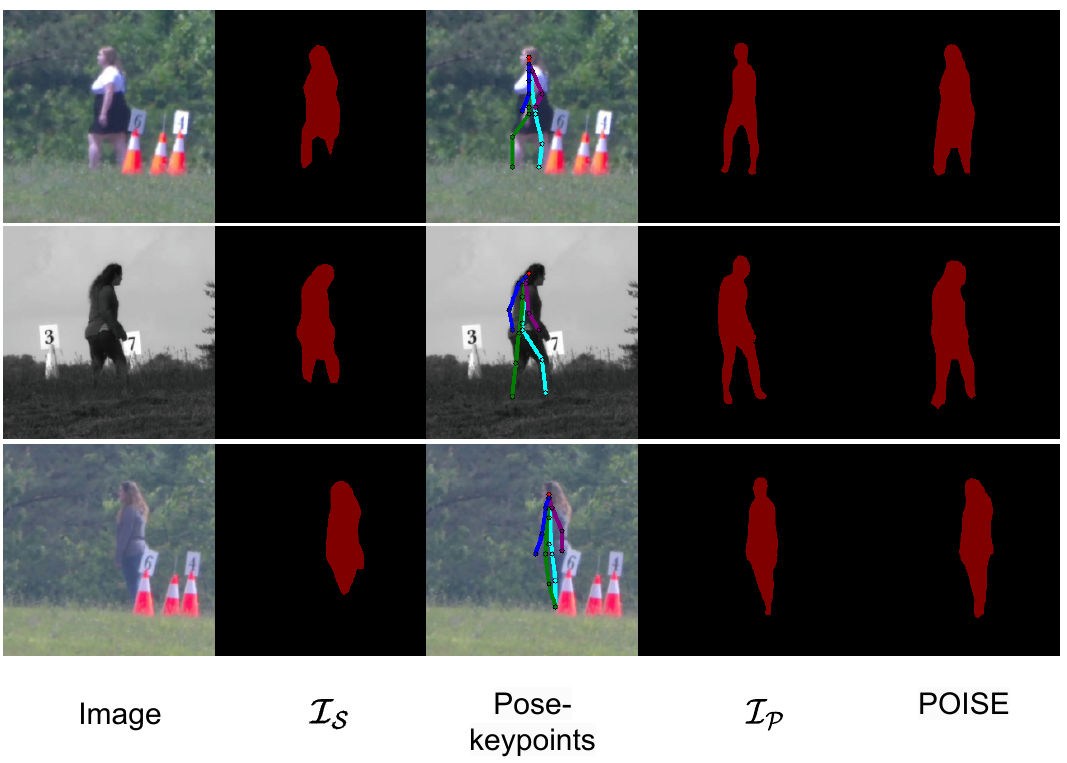}
    \caption{\textbf{Qualitative results on BRIAR.} Silhouettes extracted using \POISE compared against $\mathcal{I_{S}}$ and $\mathcal{I_{P}}$ on natural occlusions encountered in the BRIAR dataset. Note that while we use COCO occlusions for training, the model $\mathcal{M}$ generalizes seamlessly to natural occlusions.}
    \label{fig:briar}
\end{figure}

We also assess the performance of \POISE on the BRIAR dataset under natural occlusion scenarios. In Figure~\ref{fig:briar}, we visually demonstrate how \POISE excels at extracting silhouettes, closely resembling the original body shape from $\mathcal{I_{S}}$. In occluded regions, \POISE relies on $\mathcal{I_{P}}$ for guidance. Further, as shown in Figure~\ref{fig:briar}, when dealing with human subjects of diverse body shapes (compared to the synthetic dataset $\mathcal{D}$), $\mathcal{I_{P}}$ fails to retain body-specific information, while $\mathcal{I_{S}}$ retains most of it. \POISE combines information from both $\mathcal{I_{S}}$ and $\mathcal{I_{P}}$ to produce optimal silhouettes.

\begin{figure}[!htb]
    \centering
    \includegraphics[scale = 0.5]{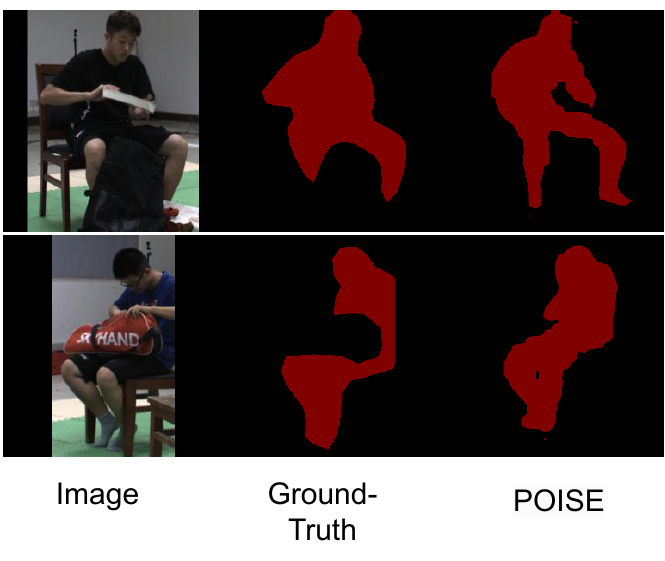}
    \caption{\textbf{Qualitative results on 3DOH50K dataset \cite{zhangoohcvpr20}} Silhouettes obtained by \POISE compared against the original ground-truth silhouettes provided. Clearly, silhouettes from \POISE are much more complete as compared against the ground-truth silhouettes.}
    \label{fig:3doh50k}
\end{figure}

Figure~\ref{fig:3doh50k} shows the efficacy of \POISE in handling natural occlusions on the 3DOH50K dataset. As \POISE uses complementary self-supervisory signals (from $\mathcal{I_{S}}$ and $\mathcal{I_{P}}$) during training, it is able to handle occlusions reasonably well. The results are particularly interesting as it shows silhouettes from an self-supervised method i.e. \POISE is better off as compared against ground-truth silhouettes. Additional results using recent segmentation methods are provided in section 4 of the supplementary.

\begin{table*}
 \caption{Average Rank-1 gait Recognition Accuracy using GaitBase \cite{fan2023opengait} across 11 different camera positions ($0^{\circ}$, $18^{\circ}$, .., $180^{\circ}$) for \POISE against $\mathcal{I_{S}}$ and $\mathcal{I_{P}}$ on the CASIA-B dataset as a function of varying {\bf occlusion duration}.}
\vskip 4 pt
\centering
\resizebox{\linewidth}{!}{
  \begin{tabular}{lllllllllllllllll}
    \toprule
    \multirow{2}{*}{Method} &
      \multicolumn{3}{c}{20\%} &
      \multicolumn{3}{c}{33\%} &
      \multicolumn{3}{c}{50\%} &
      \multicolumn{3}{c}{67\%} &
      \multicolumn{3}{c}{80\%} \\ \cmidrule(lr){2-4} \cmidrule(lr){5-7} \cmidrule(lr){8-10} \cmidrule(lr){11-13} \cmidrule(lr){14-16}
    & NM & BG & CL & NM & BG & CL & NM & BG & CL & NM & BG & CL & NM & BG & CL \\
    \midrule
     $\mathcal{I_{S}}$ & {\bf 84.44} & 65.58 & 48.68 & 82.35 & 64.42 & 47.51 & 79.75 & 61.19 & 44.58 & 75.45 & 57.38 & 40.99 & 71.34 & 54.31 & 39.97 \\
    $\mathcal{I_{P}}$ & 54.95 & 33.22 & 25.75 & 52.08 & 31.64 & 24.47 & 51.72 & 30.68 & 23.36 & 49.44 & 29.88 & 24.40 & 48.44 & 29.66 & 23.50 \\
    \POISE & 83.38 & {\bf 68.38} & {\bf 52.63} & {\bf 83.01} & {\bf 68.15} & {\bf 52.49} & {\bf 81.85} & {\bf 66.87} & {\bf 51.01} & {\bf 81.05} & {\bf 65.15} & {\bf 49.82} &  {\bf 78.96} & {\bf 63.54} & {\bf 49.17} \\
    \bottomrule
  \end{tabular}%
}
\label{tab:results-casia-1}%
\end{table*}

\begin{table*}

\caption{Average Rank-1 gait Recognition Accuracy using GaitBase \cite{fan2023opengait} across 11 different camera positions ($0^{\circ}$, $18^{\circ}$, .., $180^{\circ}$)  
for \POISE against $\mathcal{I_{S}}$ and $\mathcal{I_{P}}$ on the CASIA-B dataset as a function of varying {\bf occlusion severity}: 12, 16, .. 28. }
\vskip 4 pt
\centering
\resizebox{\linewidth}{!}{
  \begin{tabular}{lllllllllllllllll}
    \toprule
    \multirow{2}{*}{Method} &
      \multicolumn{3}{c}{12} &
      \multicolumn{3}{c}{16} &
      \multicolumn{3}{c}{20} &
      \multicolumn{3}{c}{24} &
      \multicolumn{3}{c}{28} \\ \cmidrule(lr){2-4} \cmidrule(lr){5-7} \cmidrule(lr){8-10} \cmidrule(lr){11-13} \cmidrule(lr){14-16}
    & NM & BG & CL & NM & BG & CL & NM & BG & CL & NM & BG & CL & NM & BG & CL \\
    \midrule
    $\mathcal{I_{S}}$ & 79.75 & 61.19 & 44.58 & 79.57 & 61.99 & 45.03 & 80.26 & 61.03 & 43.32 & 79.54 & 59.99 & 42.29 & 79.62 & 59.15 & 41.96 \\
    $\mathcal{I_{P}}$ & 51.72 & 30.68 & 23.36 & 48.78 & 30.21 & 22.56 & 46.75 & 28.31 & 22.03 & 44.72 & 27.12 & 21.46 & 43.23 & 25.93 & 20.75 \\
    \POISE & {\bf 81.85} & {\bf 66.87} & {\bf 51.01} & {\bf 81.25} & {\bf 65.52} & {\bf 50.08} & {\bf 80.47} & {\bf 66.32} & {\bf 51.11} & {\bf 80.05} & {\bf 64.70} & {\bf 49.57} & {\bf 80.36} & {\bf 65.73} & {\bf 49.02}  \\
    \bottomrule
  \end{tabular}%
}
\label{tab:results-casia-2}
\end{table*}

\begin{table}[!htb]
\centering
\caption{Quantitative Results for top-$k$ gait recognition accuracy for \POISE against $\mathcal{I_{S}}$ and $\mathcal{I_{P}}$ on BRIAR dataset using GaitBase \cite{fan2023opengait}.}
\begin{tabular}{@{}llll@{}}
\toprule
Method & top-1 & top-3 & top-5\\
\hline 

$\mathcal{I_{S}}$ &   8.96 & 22.39 & 31.72 \\

$\mathcal{I_{P}}$ &   14.20 & 28.60 & 38.68 \\
   
\POISE &   {\bf 16.66} & {\bf 31.69} & {\bf 40.54} \\

\bottomrule
\end{tabular}
\label{tab:gait-briar}
\end{table}

\begin{table}[t]
\centering
\caption{\textbf{Ablation study results.} Pseudo-label loss significantly improves rank-1 gait recognition accuracy on CASIA-B dataset (occlusion duration: 50\%, severity: 12). Best result is in bold and second result is underlined.}
\resizebox{\columnwidth}{!}{
\begin{tabular}{@{}llllllll@{}}
\toprule
Method & $\mathcal{L}_\mathcal{I_{S}}$ & $\mathcal{L}_\mathcal{I_{P}}$ & $\mathcal{L}_{pl}$ & {NM} & {BG} & {CL} & {Avg.} \\ \midrule
 $\mathcal{I_{S}}$ & \checkmark & & & 80.22 &  62.26  &  43.31    & 61.93       \\
$\mathcal{I_{P}}$ & & \checkmark & & 52.23 & 31.07 & 24.79 &   36.03   \\
\POISE & \checkmark & \checkmark & & 81.33 &63.89 & 54.35 &\underline{66.52}   \\
\POISE (full) &\checkmark  & \checkmark & \checkmark & 83.30 & 67.00 &  52.11    & {\bf 67.47}      \\
 \bottomrule
\end{tabular}}
\label{tab:ablation}
\vspace{-2pt}
\end{table}

\subsubsection{Gait Recognition}

In this section, we present the efficacy of \POISE on the downstream task of gait recognition on two datasets: CASIA-B and BRIAR. \\

\noindent {\bf Results on CASIA-B:} We present gait recognition results on the CASIA-B dataset for three different walking conditions: Normal (NM), Carrying Bag (BG) and Clothing (CL). We evaluate \POISE under the following two different settings. 

\noindent {\bf Duration of Occlusion.} In this setting the severity of the occlusion remains the same across frames but the number of occluded frames changes. As an example, $20\%$ occlusion means that for every video in the dataset, a temporally continuous set of frames of length $20\%$ of the total number of frames are occluded. Table~\ref{tab:results-casia-1} presents quantitative results on the performance of \POISE against $\mathcal{I_{S}}$ and $\mathcal{I_{P}}$ at fixed occlusion severity of 12. It is interesting to note that \POISE outperforms both $\mathcal{I_{S}}$ and $\mathcal{I_{P}}$ by significant margins under scenarios of temporally long occlusions.

 \noindent { \bf Severity of Occlusion.} Similar to our experimental settings for section~\ref{robust-silh-ext}, we study the effect of varying occlusion severity on gait recognition, keeping the same duration of occlusion. Table~\ref{tab:results-casia-2} presents quantitative results on the performance of \POISE against $\mathcal{I_{S}}$ and $\mathcal{I_{P}}$ at an occlusion duration of $50\%$. We note that, although with increasing severity of occlusion gait recognition performance falls for all the three listed methods, \POISE still outperforms $\mathcal{I_{S}}$ and $\mathcal{I_{P}}$ by significant margins. \\

\noindent {\bf Results on BRIAR:} Table~\ref{tab:gait-briar} shows the efficacy of \POISE in obtaining robust silhouettes that help with gait recognition on the BRIAR dataset. We obtain improvements of $\approx$ 7 \% and 8 \% in terms of top-1 and top-5 gait recognition accuracy over simply using $\mathcal{I_{S}}$. These improvements are significant as the videos in the dataset suffer from atmospheric turbulence and have instances of persistent natural occlusions. 

A thorough discussion with additional quantitative analysis on gait recognition on the two aforementioned datasets is provided in section 5 of the supplementary.
\subsubsection{Importance of Pseudo-Labeling:}
We present an ablation study on the importance of using pseudo-labels from the model $\mathcal{M}$ under training. Table~\ref{tab:ablation} shows that there is an improvement of $\approx 1\%$ owing to the use of the pseudo-label loss as described in Equation~\ref{eqn:pl-loss}. 

\section{Conclusion}
\label{sec:conclusion_and_future_works}
We present \POISE, an innovative approach for robust human silhouette extraction under occlusions. By leveraging pose estimation and self-supervised learning, \POISE effectively addresses the limitations posed by occlusions, ensuring accurate and preserved body shape representations. Unlike traditional methods, \POISE eliminates the need for expensive annotations, making it cost-effective and practical. Experimental results demonstrate the superiority of \POISE in improving silhouette extraction under occlusions. Furthermore, \POISE exhibits promising performance in gait recognition tasks, underscoring its potential impact in various applications. In summary, \POISE offers a straightforward yet valuable solution for extracting precise silhouettes in challenging scenarios, contributing to advancements in biometrics and related fields.

\noindent \textbf{Acknowledgements.} This research is partially supported by the Office of the Director of National Intelligence (ODNI), specifically through the Intelligence Advanced Research Projects Activity (IARPA), under contract number [2022-21102100007]. The views and conclusions in this research reflect those of the authors and should not be construed as officially representing the policies, whether explicitly or implicitly, of ODNI, IARPA, or the U.S. Government. Nevertheless, the U.S. Government retains the authorization to reproduce and distribute reprints for official government purposes, regardless of any copyright notices included. We further thank our colleague Yash Garg (ygarg002@ucr.edu) for helping us with additional experiments.

{\small
\bibliographystyle{ieee_fullname}
\bibliography{main.bib}
}

\end{document}